\documentclass{article}
\usepackage{arxiv}
\usepackage{listings}
\usepackage{soul}
\usepackage{url}
\usepackage[small]{caption}
\usepackage{graphicx}
\usepackage{amsthm}
\usepackage{algorithm}
\usepackage{graphicx}
\usepackage{amssymb}
\usepackage{amsmath}
\usepackage{amsfonts}
\usepackage{booktabs}
\usepackage{multirow}
\usepackage{xcolor}
\usepackage{comment}
\usepackage{breqn}
\usepackage{algpseudocode}

\begin{document}
\title{Some 
Recent advances in reasoning based on analogical proportions}

\author{{Myriam Bounhas}\\
Liwa College of Technology, Abu Dhabi, United Arab Emirates \\
\& LARODEC Laboratory, Tunis University, Tunisia\\
\texttt{myriam$_{-}$bounhas@yahoo.fr } 
\And {Henri Prade}\\
IRIT, CNRS \& University of Toulouse, France\\
\texttt{henri.prade@irit.fr}
\And {Gilles Richard}\\
IRIT, University of Toulouse, France\\
\texttt{gilles.richard@irit.fr}
}

\maketitle
\begin{abstract}
Analogical proportions compare pairs of items $(a, b)$ and $(c, d)$ in terms of their differences and similarities. They play a key role in the formalization of analogical inference. The paper first discusses how to improve analogical inference in terms of accuracy and in terms of computational cost. Then it indicates the potential of analogical proportions for explanation. Finally, it highlights the close relationship between analogical proportions and multi-valued dependencies, which reveals an unsuspected aspect of the former.
\end{abstract}


\maketitle

\section{Introduction}
Analogical proportions (AP) are statements of the form ``$a$ is to $b$ as $c$ is to $d$''. They compare the pairs of items $(a, b)$ and $(c, d)$ in terms of their differences and similarities. 
The explicit use of APs  in analogical reasoning has contributed to a renewal of its  applications, leading to many developments, especially in the last decade; see \cite{PraRicIJCAI2021} for a survey. 

However, even if much has been already done both at the theoretical and at the practical levels, the very nature of APs may not yet be fully understood and their full potential explored. 

In the following, we survey recent works on APs along three directions: 
\begin{itemize}
    \item their role in classification tasks \cite{BouPraSUM2022};
    \item their use for providing explanations \cite{LimPraRicSUM2022};
    \item their relation with multi-valued dependencies \cite{LinkPraRicSUM2022}.
\end{itemize}
{This just intends to be an introductory paper, and the reader is referred to the above references for more details on each issue.}  We first start with a background, before addressing the three questions above in turn.

\section{Background}
\label{AP}
An analogical proportion (AP) is a statement of the form ``$a$ is to $b$ as $c$ is to $d$'', linking four items $a,b,c,d$. It is denoted by $a:b::c:d$. 
APs are supposed to satisfy the following properties:
\begin{enumerate}
\item $a:b::c:d \Rightarrow c:d::a:b$ (\textit{symmetry});
\item $a:b::c:d \Rightarrow a:c::b:d$ (\textit{central permutation}).
\end{enumerate}
which along with 
$a \! :\! b \!:: \!a :\! b$ (\textit{reflexivity}), 
 are the basic AP postulates (e.g., \cite{PraRicLU2013}). 
 
 These properties mimic the behavior of arithmetic proportions (i.e., $a-b = c-d$) or geometric proportions (i.e., $\frac{a}{b}= \frac{c}{d}$) between numbers. Immediate consequences of postulates are
i) $a\!:\!a\!::\!b\!:\!b$  (\textit{identity});  ii) $a\!:\!b\!::\!c\!:\!d \Rightarrow d\!:\!b\!::\!c\!:\!a$ (\textit{extreme permutation});
iii) $a\!:\!b\!::\!c\!:\!d \Rightarrow b\!:\!a\!::\!d\!:\!c$ (\textit{internal reversal}); iv) $a:b::c:d \Rightarrow$ $d:c::b:a$ (\textit{complete reversal}).

When items $a$, $b$, $c$, $d$ are not reduced to a single attribute value, but are tuples of $n$ attribute values, APs are just defined  component-wise:
$$a:b::c:d \mbox{ iff }\forall i \in \{1,..., n\}, ~~a_i : b_i  :: c_i : d_i$$


We first consider the case of one Boolean attribute applied to each of the four items ($n=1$).
In that Boolean case,
 the following logical expression has been  proposed   for an AP \cite{MicPraECSQARU2009}: $${a : b :: c : d=}((a \wedge \neg b ) \equiv (c \wedge \neg d )) \wedge ((\neg a \wedge  b ) \equiv (\neg c \wedge  d ))$$ 
 This formula expresses that {\it $a$ differs from $b$ as $c$ differs from $d$ and  $b$ differs from $a$ as $d$ differs from $c$}. 
 It is only true  for  6 valuations, namely $0 : 0 :: 0 : 0;$ $ 1 : 1 :: 1 : 1;\ \  0  : 1 :: 0 : 1;\ \ 1 : 0 :: 1 : 0; \ \  0 : 0 :: 1 : 1;\ \ 1 : 1 :: 0 : 0$. This is the minimal Boolean model agreeing with the three postulates of an AP \cite{PraRicIJAR2018}. 
 
 Boolean APs enjoy a code independence property: 
 $a : b :: c : d \Rightarrow \neg a : \neg b :: \neg c : \neg d$. In other words, encoding truth (resp. falsity) with 1 or with 0 (resp. with 0 and 1) is just a matter of convention, and does not impact the AP.
 
 This easily extends to nominal or categorical values where $a,b,c,d$ belong to a finite attribute  domain $\mathcal{A}$. In that case, $a:b::c:d$ holds true only for the three following patterns $(a, b,  c, d) \!\in\! \{(g, g, g, g), (g, h, g, h), (g, g, h, h)\}$,
$\!g, h\in\mathcal{A}, g \neq h$.
This generalizes the Boolean case where $\mathcal{A}\!=\!\{0,  1\}$. 


In the Boolean and nominal case $n=1$, the equation $a:b::c:x$ where $x$ is unknown does not always have a solution. Indeed neither $0:1::1:x$ nor $1:0::0:x$ have a solution (since $0111$, $0110$, $1000$, $1001$ are not valid patterns for an AP).  Similarly, $g : h ::h : x$ has no solution in the nominal case when $g \neq h$. The Boolean solution exists if and only if $(a \equiv b) \vee (a \equiv c)$ is true. If the solution exists, it is unique and given by $x=c \equiv (a \equiv b)$. In the nominal case, the solution exists (and is unique) if $a=b$ (then $d=c$) or if $a =c$ (then $d=b$) \cite{PirYvon99}.\\ 

 {\it Note that, assuming that the AP $a:b::c:d$ is true, one can indeed recalculate $d$ from $a$, $b$, $c$.}\vspace{0.2 cm} 
 
 More generally, analogical inference   amounts to an analogical jump stating that if an AP holds between four items for $n$ attributes, an AP may also hold for an attribute $n+1$: 
$$\frac{\forall i \in \{1,..., n\}, ~~a_i : b_i  :: c_i : d_i \mbox{ holds}}{~~a_{n+1} : b_{n+1} :: c_{n+1} : d_{n+1} \mbox{ holds}}$$
If $a_{n+1}$, $b_{n+1}$, $c_{n+1}$ are known, this enables the prediction of $d_{n+1}$, provided that $a_{n+1} : b_{n+1} :: c_{n+1} : x$ is solvable. When attribute $n+1$ is a class label, analogical inference is the basis for analogical proportion-based classification \cite{BouPraRicIJAR2017b}.

\section{Improving analogical proportions-based inference }
There are usually many triplets $(a,b,c)$ in a data set that enable us, using the above inference scheme, to predict a value for $d_{n+1}$ from the fact that
$a : b :: c : d$ holds and from the knowledge of $a_{n+1}$, $b_{n+1}$, and $c_{n+1}$. In a classification problem, the attribute $n+1$ is the class label of the items. Since we may get different predictions for different triplets, a vote should takes place for finding the label that receives the greatest number of votes. This basic procedure has obviously a cubic complexity due to the search for triplets. 

After initial promising experiments
\cite{MicBayDelJAIR2008,BouPraRicECAI2014}, the question of understanding why and when analogical inference gives good results in classification has  been raised. Theoretical studies  
\cite{CouHugPraRicIJCAI2017} have shown  that APs-based inference never fails only in case the Boolean function underlying the classification is affine; this result has been extended to the case of nominal attributes \cite{CouceiroLMPR20}. But it is not only in the case of linearity or quasi-linearity that the analogical inference leads to good results in classification in practice. Indeed we do not need that $100\%$ of votes  be right to have a correct prediction, you just need a majority!

This situation raises several issues that may appear worth of investigation:
\begin{itemize}
    \item  Is there  a way to determine from a training set if analogical inference is appropriate for this data set?
     \item Would it be possible to modify the definition of the analogical proportion in order to get better results?
      \item Is it possible to reduce the computational complexity? and if yes, how? 
    \item Is there a way to reduce the number of triplets used to predict the class labels?
    \item Is there a way to improve analogical inference?
\end{itemize}
 In what follows, we briefly address the first three questions, before dealing with the last two in more detail.

\paragraph{Analogical suitability of a data set} Some data sets may not be suitable for the AP-based inference in the sense that its application would lead to poor results in classification. This can be tested on the training set by computing the proportion of examples for which the AP-based inference yields a false value as a prediction (each example is in turn withdrawn of the training set and its label computed on the basis of analogical inference).

\paragraph{Analogical proportions and related proportions} Boolean analogical proportions belong to the family of logical proportions, which are quaternary connectives that share a common logical structure and are all true for $6$ of the $2^4$ possible valuations  \cite{PraRicLU2013}. Among logical proportions, only $8$ are symmetrical and code independent, including AP  \cite{PraRicIFCOLOG2014}. 
Because different logical proportions have different semantics, they serve different purposes. It is not clear that any of them can be substituted for AP in the inference scheme with beneficial results (even if some of them may be used differently for classification purposes \cite{BouPraRicIJAR2017a}). We might also think of reducing \cite{PraRicIJAR2018} or enlarging \cite{Klein1983} the set of patterns that make an AP true by modifying its definition, but there is no known evidence that it could be beneficial either.

\paragraph{Complexity} It has been observed that choosing triplets where $c$ is a near neighbor of $d$ does not harm the quality of the results \cite{BouPraRicIJAR2017b}. Another way to decrease the computational cost may be  to choose the pairs $(a,b)$ in the triplets we consider. This is the topic of the next subsection.

\paragraph{Choosing triplets}
Following the very semantics of an AP, it is useful to analyse an AP in terms of pairs and in terms of attributes where the pairs agree and where they disagree.  

Given two  Boolean vectors $a$ and $b$, their \emph{agreement set} $Ag(a,b)$ is the set of attributes
$Ag(a,b)=_{def}\{i \in \{1,..., n\}\ |\ a_i =b_i \}$. 
Their \emph{disagreement set} $Dis(a,b)$ is the set of attributes
$Dis(a,b)=_{def}\{i \in \{1,..., n\} \ |\ a_i \neq b_i\}$.

{\it AP as a pair of  pairs.} In  Table \ref{APpairs}, in all generality, the attributes have been grouped and ordered in such a way that $a$, $b$, $c$, $d$ coincide on the attributes from ${A}_1$ to  $A_{j-1}$. For the attributes from $A_{j}$ to  $A_{\ell-1}$, $a$ and $b$ are equal, and so are $c$ and $d$ (but not in the same way), i.e., $Ag(a,b)=Ag(c,d) = \{1, \dots, \ell-1\}$. For the remaining attributes from $A_\ell$ to $A_{n}$, $a$ differs from $b$ in the same way as $c$ differs from $d$. The conditions $Ag(a,b)=Ag(c,d)$ and $Dis(a,b)= Dis(c,d) (= \{\ell, \dots, n\}$) are not enough  for ensuring $a:b::c:d$, which also  requires that the differences between $a$ and $b$ and between $c$ and $d$ are in the same sense. We can easily see that  $a:b::c:d$ is true on the attributes from $A_1$ to  $A_{n}$ (Table \ref{APpairs} exhibits vertically the 6 patterns that make  an AP true), and that any situation where $a:b::c:d$ is true can be described in this way. We use Boolean attributes  in  Table \ref{APpairs} for attributes  $A_1$ to $A_{n}$. Replacing some attribute values 1 with $g$ and 0 with $h$, which would be the case for nominal attributes, would not change the above analysis. 

Thus in Table \ref{APpairs}, $a$ and $b$ on the one hand, and $c$ and $d$ on the other hand are paired in the same way: the items of each pair are equal on the same attributes (${A}_1$ to ${A}_{\ell - 1}$), and $a$ and $b$ differ from each other exactly as $c$ and $d$ differ (attributes  ${A}_\ell$ to ${A}_{n}$). Observe that  
if we swap  $b$ and $c$, the APs are preserved (as expected), but now 
$a$ and $c$ (resp. $b$ and $d$) are equal on attributes ${A}_1$ to ${A}_{j - 1}$ and ${A}_{\ell}$ to ${A}_n$, while they differ on attributes ${A}_j$ to ${A}_{\ell - 1}$.

\begin{table}[!ht]
\centering
$
\begin{array}{c|cccccc|c|}
&  A_1 \mbox{-} A_{i-1} & A_i \mbox{-} A_{j-1}
& A_j \mbox{-} A_{k-1} &  A_k \mbox{-} A_{\ell-1} & A_\ell  \mbox{-} A_{m-1}& A_m   \mbox{-} {A}_{n}&\mathcal{C} \\
\hline
\!\!\! a  & \!\!\!1 \!\!\! \!\!   &  \!\!\!\!\!\! 0\!\!\!\!\!\!  &    \!\!\!\!\!\! 1\!\!\!\!\!\!  &    \!\!\!\!\!\!0\!\!\!\!\!\!  & \!\!\!\!1\!\! \!\!& 0 &  p\\
\hline
\!\!\! b  & \!\!\!1 \!\!\! \!\!   &  \!\!\!\!\!\!0 \!\!\!\!\!\!   & \!\!\!\!\!\!1\!\!\!\!\!\!   &  \!\!\!\!\!\!0\!\!\!\!\!\! & \!\!\!\!\!\!0\!\!\!\!\!\! &  1 &  q  \\
\hline
\hline
\!\!\! c & \!\!\!1 \!\!\! \!\!   &   \!\!\!\!\!\! 0\!\!\!\!\!\! &   \!\!\!\!\!\!0\!\!\!\!\!\!  &  \!\!\!\!\!\!1\!\!\!\!\!\! & \!\!\!\!\!\!1\!\!\!\!\!\! &  \!\!\!\!\!\!0\!\!\!\!\! & r  \\
\hline
\!\!\! d  &\!\!\!1 \!\!\! \!\!   &   \!\!\!\!\!\! 0\!\!\!\!\!\! &   \!\!\!\!\!\!0\!\!\!\!\!\!  &  \!\!\!\!\!\!1\!\!\!\!\!\! & \!\!\!\!\!\!0\!\!\!\!\!\! &  \!\!\!\!\!\!1\!\!\!\!\!  & s\\
\hline
\end{array}
$
\caption{AP: Pairing pairs}\label{APpairs}
\end{table}
{\it Pairs as rules.}
In Table \ref{APpairs}, suppose that the attributes from $A_1$ to $A_{n}$ are used to describe 
situations for which a conclusion, or a class label $\mathcal{C}$  is associated to them. $\mathcal{C}$ may be a Boolean or a nominal attribute. It is also assumed that the AP $p:q::r:s$ is true, which means that $p=q$ and $r=s$, or that $p=r$ and $q=s$, in column $\mathcal{C}$. This fits with the analogical prediction of $\mathcal{C}(d)$ from 
$\mathcal{C}(a)$, $\mathcal{C}(b)$, and $\mathcal{C}(c)$ when $a:b::c:d$ holds on attributes $A_1$ to $A_n$.

Let us examine the case where $p=r$ and $q=s$ with $p \neq q$ (the other case $p=q$ and $r=s$ can be obtained by exchanging $b$ and $c$). 
The tilting of the $\mathcal{C}$ value from $p$ to $q$ between $a$ and $b$ and between $c$ and $d$ can be explained, in view of the attributes considered, only by the change of values of the attributes from ${A}_\ell$ to ${A}_n$ (which is the same for pair $(a, b)$ and pair $(c, d)$).

The two pairs $(a, b)$ and $(c, d)$, which correspond to different contexts (described by the attributes ${A}_1$ to ${A}_{\ell - 1}$, and differentiated by values of attributes ${A}_j$ to ${A}_{\ell - 1}$), suggest to see these pairs as instances of a \emph{rule} expressing that 
the change on the attributes  ${A}_\ell$ to ${A}_{n}$ determines the change on $\mathcal{C}$ \emph{whatever} the context. However, this rule may have exceptions in the data set. Indeed nothing forbids that there exist items
$a'$ and $b'$ such that $a' : b' :: c : d$ holds on attributes ${A}_1$ to ${A}_{n}$ and $\mathcal{C}(a') = \mathcal{C}(b')=p$, which would lead to the analogical prediction $\mathcal{C}(d)=s = r = p$ (remember we assumed $p=r$), contradicting the fact that $s=q$  (since $p \neq q$).

So we may calculate the confidence and the support of the rule associated with pair $(a, b)$ and  pair $(c, d)$, in the data set. This reflects the \emph{competence} of the pairs  
 \cite{iccbr/LieberNP19}. 
 
 Given a new item $d$ for which the class label is unknown, in a standard $AP$-classifier, the process requires to look at all triplets $(a,b,c)$ of examples in the training set with known labels that makes an analogical proportion with $d$ and such that the class equation is solvable. This search process is brute-force and takes longer the larger the dataset. 
But, a restricted set of triplets $(a,b,c)$ is enough for correctly predicting the label for $d$, as explained now.

During the testing phase, instead of systematically considering all pairs $(a,b)$, we restrict the search for suitable triplets $(a,b,c)$ by constraining $(a, b)$ to be one of the competent pairs, previously extracted during the training phase, while $c$ is taken as any training example.
In fact, experiments show that one can further constrain  $c$ to be a near neighbor of $d$ (which means that $a$ and $b$ are also near neighbors) without any harm \cite{BouPraRicIJAR2017b}, making complexity {quadratic}.

This  has a double merit: it  first considerably simplifies the algorithm during the testing phase; although the complexity to classify new items is still {quadratic due to the use of pairs $(a,b)$, $c$ being a close neighbor}, however the number of explored triplets is very small if compared to their original number. More importantly, new items are now classified using $only$ competent pairs which means that there is less chance of making prediction errors.
Indeed in the context of case-based reasoning, a preliminary study \cite{iccbr/LieberNP19} has  shown that the use of competent pairs improves the extrapolation process when applied to a Boolean case base. This can be also observed on the Boolean or nominal datasets described in Table \ref{des}. The results given in Table \ref{OptACnom} are for 
near neighbors at distance $k=2$, using only $50\%$ of the dataset for selecting triplets; $BaselineAC$ corresponds to the brute force algorithm.

\begin{table}[ht]

 \begin{center}
 \scalebox{0.9}[0.9]{
\begin{tabular}{|l | c |c |c| c|}
  \hline
 Datasets & Instances & Nominal/Binary Att. &   Classes \\
 \hline

 Balance &   625 &  4 &3 \\
 Car  &    743 &   7 &  4  \\
 TicTacToe & 521 &   9 & 
2 \\
 Monk1 &      432 &   6   &  2 \\
 Monk2 &     432 &    6   &  2 \\
 Monk3 &      432 &    6   &  2 \\
  \hline
\end{tabular}}
\caption{Description of considered datasets} \label{des}
\end{center}
\end{table}

\begin{table}[!htbp]
\begin{center}
\scalebox{0.8}[0.8]{
\begin{tabular}{|l|c|c| }
\hline
\textbf{ Datasets} & 
\textbf{BaselineAC } & \textbf{Selected Triplets } \\

 \hline
   &   {\it Accuracy}  & {\it Accuracy}  \\

\hline
Balance  &   88.45 $\pm$ 4.05& \textbf{91.71 $\pm$ 3.43}
 \\

\hline        	    	

Car  & 88.13 $\pm$ 3.91 & \textbf{96.88 $\pm$2.01}
  \\
 \hline

 TicTacToe &  \textbf{100}&   \textbf{100} \\

 \hline
Monk1  &  99.31 $\pm$ 1.20 &  \textbf{100}
 \\
 \hline

Monk2   & 70.69 $\pm$ 7.65 &  \textbf{100}
 \\
 \hline

Monk3   &  95.28 $\pm$ 3.12 & 99.17 $\pm$ 1.56 
  \\
\hline



\hline
\end{tabular}
}
\caption{Results for a classifier based on selected triplets  for nominal datasets }\label{OptACnom}
\end{center}
\end{table}

\paragraph{Improving analogical inference}
The classification problem may rely on a logical analysis of the available examples. 
Consider two distinct examples $a$ and $b$. They coincide on a subset of attributes $Ag(a, b)$ and they differ on the subset $Dis(a,b) \neq \emptyset$. There are two cases:%

- If $\mathcal{C}(a)= \mathcal{C}(b)$, this means that at least in the context defined by the values taken on $Ag(a, b)$ (if non-empty) the difference between $a$ and $b$ observed on $Dis(a,b)$ does not affect the class.

- If $\mathcal{C}(a)\neq \mathcal{C}(b)$, it means that the change in $Dis(a,b)$ is enough
    for explaining the change from  $\mathcal{C}(a)$ to $\mathcal{C}(b)$ in the context defined by the values taken on $Ag(a, b)$ (if non empty).
To what extent what is true on a  pair may be general?

Let $d$ be a new item (differing from any example) for which $\mathcal{C}(d)$ is not known. 

This analysis leads to first consider the items that differ from $d$ in only one attribute.
{If no such items can be found, we may consider the items that differ from $d$ in two attributes (or more if necessary).}
Let $c$ be one of them.
Let  $dif(c, d) = (c_1 - d_1, \cdots, c_n - d_n) \in \{-1, 0, 1\}^n$.
Let us look at all the pairs $(a, b)$ of examples such that  $dif(a, b) = dif(c, d)$ (thus $Ag(a, b) = Ag(c, d)$) to see the effects of this difference. We have 3 cases: \begin{itemize}
  \item[case 1] $\forall (a, b)$ such that $dif(a, b) = dif(c, d)$, we have $\mathcal{C}(a) = \mathcal{C}(b)$.
  Then there is no reason  not to expect  $\mathcal{C}(d)= \mathcal{C}(c)$ according to the considered $c$;
  \item[case 2] $\forall (a, b)$ we have $\mathcal{C}(a) \neq \mathcal{C}(b)$ then we are led to predict $\mathcal{C}(d) = \mathcal{C}(b)$  according to the considered $c$ {if $\mathcal{C}(c) =\mathcal{C}(a)$}. Indeed in all pairs   that present the same difference $dif(c, d)$
  {there is a change of class and}
  there is no reason not to observe the same change  from  $c$ to $d$;
  \item[case 3] We have two non-empty sets of pairs: i) the ones such that $\mathcal{C}(a) = \mathcal{C}(b)$ and ii) the others such that $\mathcal{C}(a) \neq \mathcal{C}(b)$.
\end{itemize}

As can be seen, the two first cases are in full agreement with the analogical inference. 
The last case induces a new step in the procedure in order to understand why the same change inside the pairs lead to different classes or not for the items in the pairs according to the context.
So the problem is to look for a property  {$P$} that is true in the context of the pairs where $\mathcal{C}(a) = \mathcal{C}(b)$ and which is false for the pairs such that $\mathcal{C}(a) \neq \mathcal{C}(b)$.
This is a Bongard problem \cite{Bong} (see also \cite{Hof}, \cite{Found}), {i.e., a problem where a set has to be separated in two subsets  such that there is a property  that is true for all the elements of one subset and that is false for all the elements of the other subset.}

If the problem has a solution, let $P$ be the set of properties that separates the two subsets of pairs. {Note that $P$ is expressed in terms of attributes in $Ag(a, b) =Ag(c, d)$.}
Then if $d$ has property/ies $P$ then $\mathcal{C}(d) = \mathcal{C}(c)$ for this $c$;
otherwise $\mathcal{C}(d)= \mathcal{C}(b))$  for this $c$ (if $\mathcal{C}(c) = \mathcal{C}(a)$). If no solution $P$ can be found,  take another $c$.
More generally, we may consider all the  $c$ that differ from  $d$ in one attribute.
If necessary, we may
consider  the items $c$ that differ from $d$ on two properties,  and so on
until we succeed in making a prediction for $d$.

Thus, in the  procedure for predicting   $\mathcal{C}(d)$ we first look at
each example that is a 
{close} neighbor $c$ of $d$ , one by one. 
For each neighbor $c$, we compare the pair $(c, d)$ to other existing pairs $(a, b)$ and then select the subset of pairs $(a,b)$ such that $(a,b)$ and $(c, d)$ differ on the same attributes in the same way, i.e., $dif(a,b)=dif(c, d)$. The main idea is to estimate {how}
the difference between attributes 
 {influences} the prediction result.
As already explained three cases may be encountered.
In case 3  if no property $P$ could be found, the algorithm looks for another neighbor $c$. 
Since many neighbors $c$ may lead to different predictions, a majority vote is applied to decide for the final label for $d$.

\begin{table}[!htbp]

 \begin{center}
 \scalebox{0.9}[0.9]{
\begin{tabular}{| l| c  |c | c | c |c | c | }
  \hline\
Dataset &\multicolumn{2}{c|}{ AP + Bongard}    &  \multicolumn{2}{c|}{kNN}  \\
  \hline
     &  & $k^*$    & &  $k^*$ \\
   \hline

Balance &  \textbf{95.36 $\pm$ 2.59 } & 7  & 83.94 $\pm$ 4.23  & 11  \\\hline

Car  & \textbf{95.33 $\pm$ 2.40 } &3 & 92.33 $\pm$3.10  & 1   \\\hline

TicTacToe & \textbf{100} & 1  &98.27 $\pm$ 1.77  & 1  \\\hline

Monk1 &  \textbf{100}& 1  & 99.95 $\pm$ 0.  & 3   \\\hline

Monk2 & \textbf{100}& 1  	& 64.44 $\pm$ 6.99 & 11  \\\hline

Monk3 & \textbf{100}& 1  & \textbf{100} & 1  \\\hline

\end{tabular}
}
\caption{Accuracy results (means and standard deviations) for the AP + Bongard method}  \label{res2}. 
\end{center}
\vspace{-0.4cm}
\end{table}

Results for the  AP + Bongard method, obtained on the datasets considered previously,  are reported in Table  \ref{res2} (for optimized values of $k$). They look competitive  with those of Table \ref{OptACnom}. This has to be confirmed by more extensive experiments.

Thus, we have proposed two quite different options for avoiding the computational burden of the brute force approaches and improving its results: i) selecting competent pairs and then triplets, or ii) developing a logical analysis of the examples that involved both analogical proportions and the solving of Bongard problems, which are related to another logical proportion, as recalled now.

 It is worth saying that \emph{Bongard problems} are related to a logical proportion called ``inverse paralogy'', which is also a symmetrical, code-independent,  quaternary
logical connective that expresses that ``what $a$ and $b$ have in common, $c$ and $d$ do not have in common, and vice versa''. 
It
 is defined componentwise by $$IP(a, b, c, d)= [(a \wedge b) \equiv (\neg c \wedge \neg d)]\wedge  [(\neg a \wedge \neg b) \equiv ( c \wedge d)].$$ $IP(a, b, c, d)$ is true only for  6 valuations, namely $(a, b, c, d) \in\{ (1, 1, 0, 0), (0, 0,1,1),$ $ (0, 1, 1, 0),(1, 0, 0, 1),$ $ (1, 0, 1, 0), (0, 1, 0, 1)\}$, otherwise it is false.
See \cite{ipmu/PradeR16} for details on how solving Bongard problems using IPs.

\section{Analogical proportions-based explanations}
Huellermeier  \cite{HullermeierMDAI20} was the first to recently write a plea for the explanatory use of APs in 
classification and preference learning. We briefly investigate this idea in the following.

Let us simplify Table \ref{APpairs} into Table \ref{exp} where we describe in a simplified way  what is an AP in the nominal case. Moreover, we have singled out a (nominal) attribute called $Result$, 
supposed to depend on the other attributes, it may be the class to which the tuple belongs, or the result of an evaluation / selection for each tuple. We have then identified in Table \ref{exp} the roles played by each subset of attributes. Note that $X,Y,Z$ can be seen as distinct subsets of indices such that 
$X \cup Y \cup Z=R$.
\begin{itemize}
    \item a subset of attributes $X$ having the  same values for the four tuples (Boolean example: $0 0  0 0$)
    \item a subset of attributes $Y$ stating the different contexts of pairs $(a, b)$ and $(c, d)$ (Boolean example: $0 0 1 1$)
    \item a subset of attributes $Z$ describing the change(s) inside the pairs (Boolean example: $0 1 0 1$), which may be associated or not with a change on the value of $Result$. 
\end{itemize}  

\begin{table}[!ht]
\centering
$
\begin{array}{c||c|c|c||c|}

&  X \mbox{ (shared values)}  &  Y\mbox{ (context)} &Z\mbox{ (change)} &\mbox{Result} \\
\hline
a  & \!\!\!s \!\!\! \!\!     &    \!\!\!\!\!\! t\!\!\!\!\!\!   & v &  p\\
\hline
b & \!\!\!s \!\!\! \!\!     &    \!\!\!\!\!\! t\!\!\!\!\!\!   & w &  q\\
\hline
c  & \!\!\!s \!\!\! \!\!     &    \!\!\!\!\!\! u\!\!\!\!\!\!   & v &  p\\
\hline
d  & \!\!\!s \!\!\! \!\!     &    \!\!\!\!\!\! u\!\!\!\!\!\!   & w &  ?\\
\hline

\end{array}
$
\caption{How an AP looks in the nominal case}\label{exp}
\end{table}

We recognize the schema of analogical inference in Table \ref{exp}: $p: q :: p : x$ always has a unique solution $x=q$ (as explained at the end of Section \ref{AP}) and $q$ is the predicted result value for item $d$. The case $p: p :: q : x$ can be obtained by central permutation of $b$ and $c$, exchanging $Y$ and $Z$. 

Table \ref{exp} can also be seen as a basis for presenting analogical proportion-based explanations \cite{PraRicAFIA22}. Indeed the answer to the question ``why $Result(d)$ is not $p$?'' is to be found in the values taken by the subset $Z$ of $change$ attributes for $d$. Note that when $c$ is a close neighbor of $d$,  the size of $Z$ is small. This looks like  the  definition of a {\it contrastive} explanation \cite{icml/0001GCIN21} related to $d$, namely we have


\quad\quad$$\exists x \in r   \ [  Result(x)=p \wedge \bigwedge_{j \in X \cup Y} (x_j=d_j) \wedge (Result(d) \neq p) ]$$

\noindent where $r$ denotes the relational table made of the union of the tuples in the set of examples. Such an $x$ (in our example this is $c$) could  be termed as an {\it adverse example}. It is not necessarily unique. But the analogy-based  explanation is richer because we know a pair (here $(a, b)$), with another $context$ value, where the same change of attribute values leads to the same change in $Result$  as in pair $(c, d)$, which suggests the possibility of the following rule (with an abductive  flavor): 


\quad\quad$$\forall \ s, t,
(context=t) \wedge    (change = w) \to Result((s,t, w)) = q$$

However, nothing forbids that $\exists$ $(a',b') \in r^2$  such that $a'=(s, t', v)$, $b'=(s, t', w)$ with $Result(a')= Result(b')=p$, which would provide  an exception to this rule. The strength of the explanation  would depend on the relative cardinalities of pairs such as $(a,b)$ and $(a',b')$. We are back to the idea of competent pairs.

Let us illustrate the kinds of AP-based explanation with an example of a decision with multiple options \cite{BilPraRicWilFQAS2017}. The decision is whether to serve a coffee with or without sugar, with or without milk  to a person in a medical facility. As shown in the table below, 
 in situation $sit_1$ with contraindication ($contraind.$), it is recommended to serve coffee alone, in situation $sit_1$ without $contraind.$, coffee with sugar no milk, while in situation $sit_2$ with $contraind.$ we serve coffee with milk only. What can be done in $sit_2$ without $contraind.$ ? Common sense suggests coffee with sugar and milk. This is what the analogical inference gives: in fact $coffee:coffee::coffee:x$, $no:yes::no:y$ and $no:no::yes:z$ have as solutions $(x, \ y,\ z) = (coffee,yes, yes)$, as in  Table \ref{examplepo}. 
 \begin{table}[!ht]
\centering
$\begin{array}{c||c|c||c|c|c|c|}
case & situation	&	contraind.  & dec. & 	with \ sugar & with \ milk  \\
\hline
 a & sit_1 & yes & coffee & no & no \\
\hline
 b & sit_1 & no	& coffee & yes & no \\
\hline
c & sit_2 & yes &	coffee &	 no &  yes \\
\hline
\hline
d & sit_2 & no	& {\bf coffee}	 & \mathbf{yes} & \mathbf{yes} \\
\hline
\end{array}
$
\caption{An illustrative example}
\label{examplepo}
\end{table}

Thus, to the question ``why milk and sugar for $d$?", we can answer ``because we are in $sit_2$ (and not in  $sit_1$)" for milk, and ``because there is no $contraind.$'' for sugar. To the question ``why no milk for $b$?", we get the answer ``because we are in $sit_1$ and not in $sit_2$". As can be seen, the switching of $with \ sugar$ (resp. $with \ milk$) from $no$ to $yes$ is associated with the change of $contraind.$ from $yes$ to $no$, (resp. the change of $situation$ from $sit_1$ to $sit_2$). 
This example suggests that APs have an  explanatory potential from data, for answering ``why'' and also ``why not'' questions. 

This process applies to any
categorical data sets and is classifier agnostic. 
In practice, for providing an explanation we have first to determine what are the relevant attributes (using methods such as chi-square, mutual information, etc.). Then for explaining why $Result$ has not the expected positive value in some case, we have to look for contrastive examples $c$ which are close to case $d$ in terms of relevant attributes, but for which $Result$ is positive. 
Then finding other pairs $(a, b)$ exhibiting the same attribute change and the same tilting of $Result$, provided that they are competent enough, will strengthen the explanation.

\section{Analogical proportions and multivalued dependency}
We finally point out an unsuspected connection between analogical proportions and multivalued dependencies in databases. First, it is easy to build AP examples with a database flavor, as in 
Table \ref{Example}  with  nominal attributes.

\begin{table}[!ht]

\begin{tabular}{c|ccc|}
\hline
&\emph{course} & \emph{teacher}  & \emph{time}  \\
\hline
  a &Maths & Peter & 8 am \\
 b & Maths & Peter & 2 pm  \\
 c & Maths & Mary & 8 am   \\
\hline
 d & Maths& Mary  & 2 pm  \\ 
\hline
\end{tabular} 
\caption{AP: example with  nominal attributes}
\label{Example}
\end{table}

Moreover if we go back to Table \ref{exp}, a database reader may notice 
that \begin{itemize}
    \item [i)] if $p=q$, a case left aside, it would  suggest that the rule $X=s \to Result = p$ may hold, and even that a functional dependency $X \to Result$ might hold;
    \item [ii)]using  notations in Table \ref{exp}, if for all $a, b, c$ in a relational  table $r$ there exists $d$ in $r$, the weak multivalued dependencies $X \twoheadrightarrow_w Y$ and $X \twoheadrightarrow_w Z$ hold in $r$;
    \item [iii)] if for all $a, d$ in $r$ there exist $b, c$ in $r$, the  multivalued dependencies $X \twoheadrightarrow Y$  and $X \twoheadrightarrow Z$ hold in $r$. It is also true changing $Z$ into  $Result$.
\end{itemize}

Let us explain why it is so, and let us recall what are (weak) multivalued dependencies. We use standard database notations. Let $R$ be a relation  schema viewed as a set of attributes; $X$ and $Y$ denote subsets of attributes. A tuple $t$ is a complete instantiation of the attributes in $R$ describing some existing item. A relation $r$ over $R$ is a finite set of tuples over $R$. The restriction of a tuple $t$ to the attributes in $X \subseteq R$ is denoted by $t[X]$. $t[XY]$ is short for $t[X\cup Y]$.

Functional dependencies and multivalued dependencies play an important role in the design of databases. A functional dependency $X \rightarrow Y$ ($X \subseteq R$ and $Y \subseteq R$) states that for any pair of tuples $t_1$ and $t_2$ obeying the relational schema $R$, if $t_{1}[X]=t_{2}[X]$ then $t_{1}[Y]=t_{2}[Y]$, which reads ``$X$ determines $Y$''.
 
Departing from a functional dependency, the definition of a multivalued dependency requires the existence of particular tuples in the data base, under some conditions: The {\it multivalued dependency} \cite{tods/Fagin77,sigmod/BeeriFH77} (see also \cite{fuin/HartmannL09})
   $X \twoheadrightarrow Y$ 
    (which can be read as ``$X$ multidetermines  $Y$'') holds on $R$ if,  for all pairs of tuples $t_{1}$ and $t_{2}$ in $r$ such that $t_{1}[X]=t_{2}[X]$, there exists some tuple $t_3$ in $r$ such that
    $t_{3}[XY]=t_{1}[XY]$ and $t_{3}[X(R\setminus Y)]=t_{2}[X(R\setminus Y)]$. Note that, as a consequence of the definition there also exists a tuple $t_4$ in $r$ such as $t_{4}[XY]=t_{2}[XY]$ and $t_{4}[X(R\setminus Y)]=t_{1}[X(R\setminus Y)]$ (swapping the roles of $t_{1}$ and $t_{2}$).
    
    Thus altogether, when $X \twoheadrightarrow Y$ holds, for all pairs of tuples $t_{1}$ and $t_{2}$ in $r$ such that $t_{1}[X]=t_{2}[X]$, there exist tuples $t_{3}$ and $t_{4}$ in $r$ such that 
    \begin{itemize}
        \item  $t_{1}[X]=t_{2}[X ]=t_{3}[X ]=t_{4}[X]$
      \item $t_{1}[Y ]=t_{3}[Y]$
      \item $t_{2}[Y]=t_{4}[Y]$
      \item $t_{1}[R\setminus (X \cup Y )]=t_{4}[R\setminus (X \cup Y )]$
       \item $t_{2}[R\setminus (X \cup Y )]=t_{3}[R\setminus (X \cup Y )]$
\end{itemize}

A more simple, equivalent version of the above conditions can be expressed as follows: if we denote by $(x,y,z)$ the tuple having values $x$, $y$, $z$ for subsets $X$, $Y$, $R\setminus (X \cup Y)$ respectively, then whenever the tuples $(p,q,r)$ and $(p,s,u)$ exist in $r$, the tuples $(p,q,u)$ and $(p,s,r)$ should also exist in $r$. 
Note that in the definition of $X \twoheadrightarrow Y$, not only the attributes in $X$ and in $Y$ are involved, but also those in $R\setminus (X \cup Y)$, which departs from functional dependencies (where only the attributes in $X$ and in $Y$ are involved).

 A multivalued dependency $X \twoheadrightarrow Y$ is trivial if $Y$ is a subset of $X$, or if $X\cup Y$ is the whole set of attributes of the relation (then $R\setminus (X \cup Y)$ is empty). 
 
 In Table \ref{Example2}, the two multivalued dependencies  \{course\} $\twoheadrightarrow$ \{teacher\} and \{course\} $\twoheadrightarrow$  \{time\} hold, as can be checked.

\begin{table}[!ht]
\centering
\begin{tabular}{|ccc|}
\hline
\emph{course} & \emph{teacher}  & \emph{time}  \\
\hline
  Maths & Peter & 8 am \\
  Maths & Peter & 2 pm  \\
  Maths & Mary & 8 am   \\
 Maths& Mary  & 2 pm  \\ 
 Maths & Paul & 8 am \\
  Maths & Paul & 2 pm  \\
  Comp. Sci. & Peter & 8 am   \\
 Comp. Sci. & Mary  & 8 am  \\ 
\hline
\end{tabular} 
\caption{Multivalued dependencies: \{course\} $\twoheadrightarrow$ \{teacher\}; \{course\} $\twoheadrightarrow$  \{time\} }
\label{Example2}
\end{table}
Note that Table \ref{Example2} can be rewritten more compactly as in 
Table \ref{Example3}. 
\begin{table}[!ht]
\centering
\begin{tabular}{|ccc|}
\hline
\emph{course} & \emph{teacher}  & \emph{time}  \\
\hline
  Maths & \{Peter, Mary, Paul \} & \{8 am, 2 pm \} \\
  Comp. Sci.  & \{Peter, Mary\} & \{8 am\}   \\
\hline
\end{tabular} 
\caption{Compact writing of Table \ref{Example2} }
\label{Example3}
\end{table}

This acknowledges the fact that 
$r= \{ Maths \} \times \{ Peter, Mary, Paul \} \times \{ 8 am, 2 pm \}  \cup \{ Comp. Sci. \} \times \{ Peter, Mary \} \times \{ 8 am \}$. As can be seen, the teachers attached to the course and the time  attached to the course are \emph{logically independent} of each other.  

Indeed, a multivalued dependency exists in a relation when there are at least three attributes, say $X$, $Y$ and $Z$,  and for a value of $X$ there is a defined set of values of $Y$ and a  defined set of values of $Z$, such that the set of values of $Y$ is {\it independent} of set $Z$ and vice versa. 

Moreover the following properties hold:

If $X \rightarrow Y$, then  $X \twoheadrightarrow Y$.

    If $X \twoheadrightarrow Y$, then $X \twoheadrightarrow R \setminus Y$
    
    If $X \twoheadrightarrow Y$ and $Z \subseteq U$,  then $XU \twoheadrightarrow YZ$ 
    
    If $X \twoheadrightarrow Y$ and $Y \twoheadrightarrow Z$, then $X \twoheadrightarrow Z \setminus Y$.
    
  Multivalued dependencies are of interest in databases  since  decomposition of a relation $R$ into $(X, Y)$ and $(X, R \setminus Y)$ is a lossless-join decomposition if and only if $X \twoheadrightarrow  Y$ holds in $R$. Multivalued dependencies are involved in the 4th normal form in database normalization.

A multivalued dependency, given two particular tuples, requires the existence of other tuples. Its weak form only requires the existence of one tuple given three particular tuples. 
A {\it weak multivalued dependency} \cite{FischerG84} $X \twoheadrightarrow_w Y$ holds on $R$ if,  for all tuples $t_{1}$, $t_{2}$, $t_{3}$  in $r$ such that $t_{1}[XY]=t_{2}[XY]$ and $t_{1}[X(R\setminus Y)]=t_{3}[X(R\setminus Y)]$ there is some tuple 
$t_{4}$ in $r$ such that $t_{4}[XY]=t_{3}[XY]$ and $t_{4}[X(R\setminus Y)]=t_{2}[X(R\setminus Y)]$. It can be checked that if $X \twoheadrightarrow Y$ then $X \twoheadrightarrow_w Y$. The existence of  $X \twoheadrightarrow_w Y\ / \ Z$  is sufficient for ensuring the commutativity of $Y$ and $Z$ in the nesting process that enables us to rewrite  Table \ref{Example2} into Table \ref{Example3} \cite{fuin/HartmannL09}.

Let us go back to analogical proportions. As for multivalued dependencies they involved four tuples, which are taken by pairs. Let us consider the Boolean case first.  When one considers a pair of tuples $(a,b)$, one can distinguish between the attributes where the two tuples are equal and  the attributes where the two tuples disagree, as already emphasized. If we take two pairs $(a,b)$ and $(c,d)$ whose tuples are equal on the same attributes and which disagree {\it in the same way} on the other attributes (i.e. i.e., when $(a_i,b_i)=(1, 0)$ (resp. $(0,1)$), $(c_i,d_i)=(1, 0)$ (resp. $(0,1)$)) where $i$ refers to a particular attribute, these two pairs form an AP; see Table \ref{APpairs}.

This can be easily generalized to nominal attributes, as shown in Table \ref{exp}, where $a, b, c, d$ are equal on the subset of attributes $X$, where $a= b \neq c= d$ on the subset of attributes $Y$, and where the same change take place between $a$ and $b$ and between $c$ and $d$ for attributes in $Z$. Note that by central permutation, we can exchange the roles of $Y$ and $Z$.

Let us now first examine the {\it weak} multivalued dependency: 
 for all tuples $t_{1}$, $t_{2}$, $t_{3}$ in $r$  such that $t_{1}[XY]=t_{2}[XY]$ and $t_{1}[X(R\setminus Y)]=t_{3}[X(R\setminus Y)]$ there is some tuple 
$t_{4}$ in $r$ such that $t_{4}[XY]=t_{3}[XY]$ and $t_{4}[X(R\setminus Y)]=t_{2}[X(R\setminus Y)]$. 

Then if $t_{1}[XY]=t_{2}[XY]= (s, t)$ and $t_{1}[X(R\setminus Y)]=t_{3}[X(R\setminus Y)]= (s, v)$ there exists a tuple $t_{4}$ in $r$ such that $t_{4}[XY]=t_{3}[XY] = (s, u)$ and $t_{4}[X(R\setminus Y)]=t_{2}[X(R\setminus Y)]= (s, w)$. We recognize Table \ref{exp} with 
$t_{1}=a$, $t_{2}=b$, $t_{3}=c$, $t_{4}=d$. 
Thus there is a perfect match between a weak multi-valued dependency and an analogical proportion. In fact, the existence of $t_{4}$ in $r$ amounts to the existence of a (unique) solution for $a : b :: c : x$ in Table \ref{exp}.

The case of a multi-valued dependency is slightly different, as we are going to see. Indeed $X \twoheadrightarrow Y$ holds as soon as 
whenever the tuples $(p,q,r)$ and $(p,s,u)$ exist in $r$ on subsets $X$, $Y$, $R\setminus (X \cup Y)$, the tuples $(p,q,u)$ and $(p,s,r)$ also exist
in $r$ on subsets $X$, $Y$, $R\setminus (X \cup Y)$. This corresponds to Table \ref{fail} where (r, u, u, r) is not a valid valuation for an AP. 
So $t_{1} : t_{2} :: t_{3} : t_{4}$ does not hold (in fact, this corresponds to another logical proportion called ``paralogy'' \cite{PraRicLU2013}). Fortunately, it is possible to reorder the tuples for obtaining a valid AP. Indeed $t_{1} : t_{4} :: t_{3} : t_{2}$ does hold. This is also to be related to the fact that when two tuples (here $t_{1}$ and $t_{2}$) differ on at  least two attributes, one can always build two  distinct "intermediary" tuples in order to get an AP 
\cite{CouHugPraRicIJCAI2018}.
\begin{table}[!ht]
\centering
$
\begin{array}{c||c|c|c|}
& X & Y & R\setminus (X \cup Y) \\
\hline t_1 & p &  q & r \\
\hline  t_2 & p &s & u\\
\hline t_3 & p & q & u\\
\hline  t_4 & p &s&r\\
\hline

\end{array}
$
\caption{Multivalued dependency and the failure of the AP }\label{fail}
\end{table}

\section{Concluding remarks}
The paper presents ongoing researches dealing with analogical proportions, i) providing a discussion on ways of improving analogical classifiers, ii) showing how analogical proportions can contribute to explanations, and finally iii) how analogical proportions are linked to database concerns pointing out their relation to logical independence. 

All these works require further developments and experiments. Beyond that, other issues have to be explored such as the extension of these works to multi-valued APs in case of numerical attributes \cite{DubPraRicFSS2016}, or to multiple APs \cite{aicom/PradeR21} acknowledging the fact that APs should not be thought of in isolation, as it is often beneficial to exploit them jointly.


\end{document}